\documentclass[runningheads]{llncs}
\usepackage[T1]{fontenc}
\usepackage[utf8]{inputenc}

\usepackage{booktabs}
\usepackage{makecell} 
\usepackage{amsfonts} 

\usepackage{amssymb}
\usepackage{graphicx}
\usepackage{tikz}
\usepackage{amsmath}
\usepackage{multirow}
\usepackage{algorithm}
\usepackage{amsmath}
\usepackage{algorithmicx}
\usepackage{bbding}
\usepackage{hyperref}
 \hypersetup{
     colorlinks=true, 
     citecolor=green  
 }
\usepackage{color}

\usepackage{subcaption}
\usepackage{bbm}
\usepackage[table,xcdraw]{xcolor}
\usepackage{array}

\newcommand{\inlinesec}[1]{{\vspace{0.5em}\noindent\textbf{#1}}\hspace{0.5em}}

\newcommand{\ours}{\textsc{AnaUS}}   
\newcommand{\med}{\textsc{LP-SAM}}    

\usepackage{makecell}

\begin{document}
%
\title{Anatomy-Anchored Self-Supervision: Distilling Vision Foundation Models for Invariant Ultrasound Representation}
\titlerunning{Anatomy-Anchored Ultrasound Self-Supervision}

%
%

\author{
  Chunzheng Zhu\inst{1}
  \and Yijun Wang\inst{1}
  \and Jianxin Lin\inst{1}\thanks{Corresponding author.}
  \and Feng Wang\inst{1}
  \and Hongwei Wang\inst{1}
  \and Lei Zhao\inst{1}
  \and Shengli Li\inst{2}
  \and Kenli Li\inst{1}
}

\authorrunning{C. Zhu et al.}
\titlerunning{Anatomy-Anchored Ultrasound Self-Supervision}
\institute{Hunan University, Changsha, China\\
\email{linjianxin@hnu.edu.cn}
\and
Shenzhen Maternity and Child Healthcare Hospital, Shenzhen, China\\
}

%
\maketitle              

\begin{abstract}
Self-supervised pre-training paradigm has gained increasing prominence  for learning transferable representations in medical imaging, yet existing methods for ultrasound (US) images operate at the image or frame level, overlooking the anatomical context for clinical-aligned representation learning.
In this work, we propose an \text{A}natomy-\text{A}nchored \text{U}ltra\text{S}ound Self-Supervision framework (\text{\ours{}}) that shifts representation learning from generic visual regions to clinically meaningful anatomical structures.
Utilizing a learnable latent prompt engine alongside a one-time domain  adaptation on existing public image--mask pairs, we empower the \med{} module to achieve annotation-free anatomy delineation at scale.
Building upon this anatomical grounding, we propose a dual-policy self-supervised learning paradigm consisting of inter-view semantics-aware anatomy-separating alignment and contextual core-region prediction to enhance representation learning. Specifically, the former enforces feature invariance within identical anatomical regions while promoting discriminability across distinct structures; the latter compels the model to reconstruct corrupted regions, thereby capturing fine-grained structural details. Extensive evaluations on six public datasets demonstrate that \ours{} consistently outstrips current state-of-the-art methods while maintaining the computational efficiency essential for clinical deployment.
Code is available at \url{https://github.com/zhcz328/ANAUS}.

\keywords{Self-supervised Learning \and Ultrasound Imaging \and Anatomy-level Contrast \and Foundation Model Adaptation.}
\end{abstract}

\section{Introduction}

Ultrasound (US) imaging is a cornerstone of clinical diagnostics owing to its real-time capability, portability, and non-ionizing nature~\cite{born2021accelerating,zhu2024advancing}.
Deep neural networks have demonstrated considerable promise in automating US image analysis~\cite{gao2016describing,chen2021uscl,jiang2025mcbl}, yet their data-hungry training regimes remain at odds with the limited availability of annotated clinical data.
Self-supervised learning (SSL) addresses this bottleneck by deriving supervision from the data itself, enabling pre-training on large unlabeled corpora followed by efficient fine-tuning~\cite{he2020momentum,chen2020simple,grill2020bootstrap}.

Current SSL strategies for ultrasound broadly fall into two categories.
\textit{Frame-level} approaches leverage temporal coherence in US video to construct positive pairs~\cite{chen2021uscl,basu2022unsupervised,vanberlo2024intra}, yet they treat each frame holistically and risk encoding background artifacts and speckle noise rather than diagnostically relevant anatomy.
\textit{Region-level} methods borrowed from natural images~\cite{henaff2021efficient,wen2022self} attempt finer-grained learning, but their reliance on generic segmentation (\textit{e.g.}, K-means clustering or image-computable proposals) fails to isolate meaningful anatomical structures in US images, where boundaries are characteristically low-contrast and speckle-corrupted~\cite{biradar2015speckle}.
Neither paradigm aligns with clinical reasoning, which is inherently \textit{anatomy-centric}: diagnostic interpretation focuses on the morphology, texture, and spatial relationships of specific anatomical landmarks.


To ensure congruence with clinical reasoning, the US representation learning paradigm must prioritize anatomical structures over stochastic image patches as fundamental contrastive units. \textit{Attaining such structural awareness requires reconciling the inherent domain disparity between natural and medical imagery.} To this end, our proposed $\ours$ performs self-supervised pre-training directly at the anatomy level. This approach is powered by $\med$, a domain-specialized module that utilizes a learnable latent prompt engine and lightweight fine-tuning to bridge the domain gap, thereby achieving automated anatomy discovery.

Utilizing this delineation capability, \ours{} integrates two synergistic learning mechanisms. The first, view-invariant anatomical contrastive matching, regularizes the representation space to ensure invariance for the same structure under diverse views while maximizing separation between distinct anatomical entities. Such a formulation introduces explicit inter-anatomy discriminability, a property typically absent in generic contrastive frameworks. Complementing this, an anatomical contextual prediction task targets the reconstruction of corrupted core regions, compelling the encoder to internalize both local textural intricacies and global structural coherence. These dual objectives operate across a multi-scale feature hierarchy, ensuring that the learned representations encapsulate anatomical traits at hierarchical granularities.

\begin{figure}[t]
    \centering
    \includegraphics[width=1\linewidth]{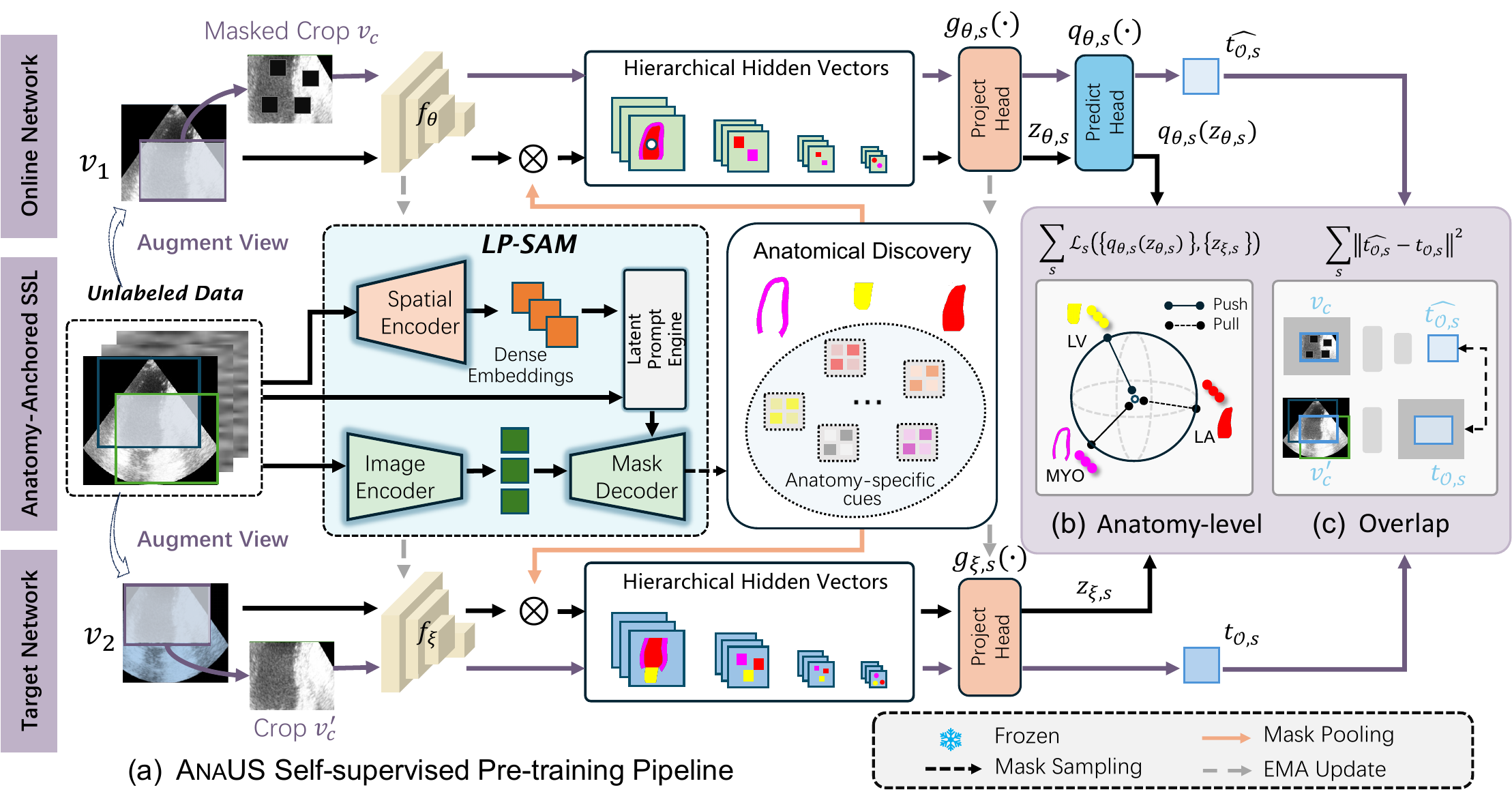}
\caption{Overview of the \ours{} framework. \textit{(i)} \med{} provides anatomical masks $\mathcal{M}(x){=}\{m_k\}_{k=1}^{K}$ for unlabeled US images via a learnable latent prompt engine (Sec.~\ref{sec:latsam}). \textit{(ii)} The encoder $f_\theta$ is jointly optimized by an anatomy-level contrastive learning objective across augmented view pairs $(\mathbf{v}_1, \mathbf{v}_2)$ (Sec.~\ref{sec:alc}) and a contextual prediction objective over corrupted core regions (Sec.~\ref{sec:cep}).}
\vspace{-1mm}
    
    \label{fig:overview}
\end{figure}

Comprehensive evaluations across six public ultrasound benchmarks, encompassing lung, breast, thyroid, and cardiac imaging, validate the multi-task efficacy of \ours{} in various downstream clinical scenarios.
The proposed framework sets a new state-of-the-art, consistently surpassing both general-purpose SSL methods and previous ultrasound-specific pre-training approaches. As a result, our work underscores the pivotal role of anatomically meaningful priors in guiding the representation learning process for medical imagery.

Our contributions can be summarized as:
\textbf{\textit{(i)}} We propose \ours{}, the first anatomy-level self-supervised pre-training framework designed explicitly for ultrasound imaging, shifting the contrastive unit from generic regions to clinically relevant structures.
\textbf{\textit{(ii)}} We introduce \med{} with a learnable latent prompt engine that empowers fully automated, expert-free structural discovery in ultrasound imagery through targeted latent-space adaptation.
\textbf{\textit{(iii)}} We demonstrate consistent state-of-the-art performance across diverse US datasets and tasks, establishing a new benchmark for ultrasound self-supervised pre-training.

\vspace{-1mm}
\section{Method}
Let $\mathcal{X}{=}\{x_i\}_{i=1}^{N}$ be a corpus of unlabeled US images. \ours{} (Fig.~\ref{fig:overview}) seeks an encoder $f_\theta{:}\,\mathcal{X}{\to}\mathcal{F}$ whose representations are invariant within anatomical structures yet discriminative across them, without any quality or segmentation annotations. Given $x$, \med{} produces a per-image mask set $\mathcal{M}(x){=}\{m_k\}_{k=1}^{K}$, $m_k \in \{0,1\}^{H \times W}$. Two augmented views $\mathbf{v}_1{=}t(x)$ and $\mathbf{v}_2{=}t'(x)$, $t,t'{\sim}\mathcal{T}$, are fed to an asymmetric online--target encoder pair optimized jointly for representation.


\begin{figure}[t]
    \centering
    \begin{subfigure}[t]{0.475\linewidth}
        \centering
        \includegraphics[width=\linewidth]{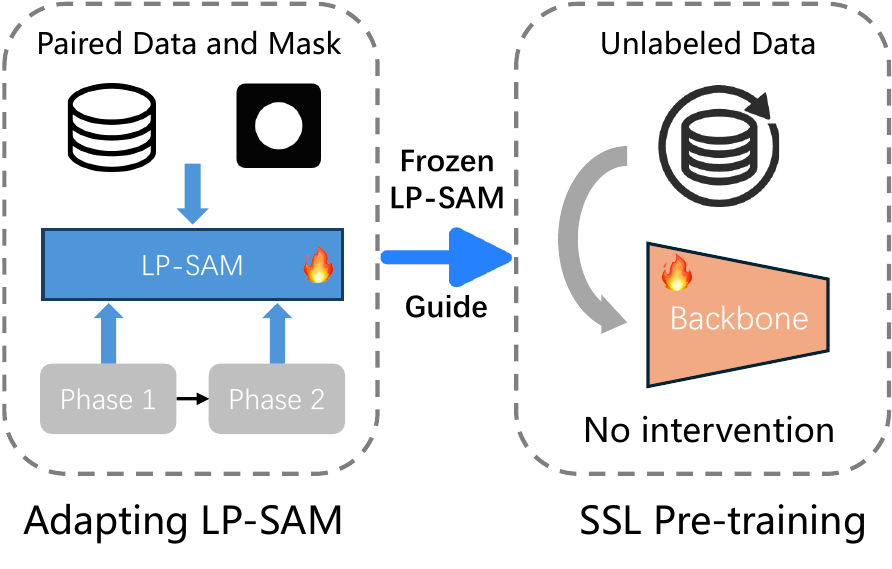}
    \end{subfigure}
    \hfill
    \begin{tikzpicture}
        \draw[dashed, gray] (0,0) -- (0,4.2cm);
    \end{tikzpicture}
    \hfill
    \begin{subfigure}[t]{0.48\linewidth}
        \centering
        \includegraphics[width=\linewidth]{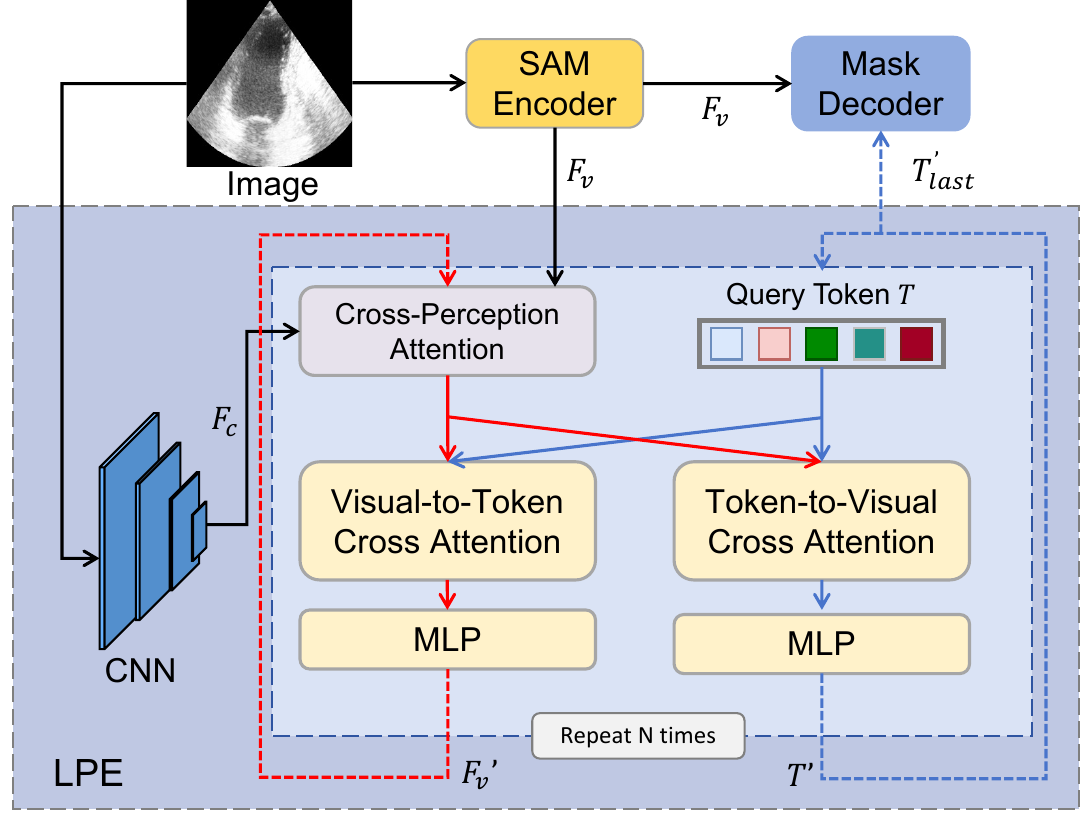}
    \end{subfigure}
 \caption{\textit{(i) Left}: The overall training pipeline of \ours{}. Phase~1 fine-tunes LP-SAM adapters on paired US image--mask data; Phase~2 trains the latent prompt engine (LPE) with LP-SAM frozen. The resulting \med{} is then frozen to guide SSL pre-training on unlabeled data. \textit{(ii) Right}: The LP-SAM architecture. In LPE, Cross-Perception Attention (CPA) fuses global ViT and local CNN features; bidirectional cross-attention iteratively refines query tokens into latent prompt embedding $T'_\mathrm{last}$ for annotation-free mask generation.}
        \label{fig:latent_prompt_engine}
\end{figure}
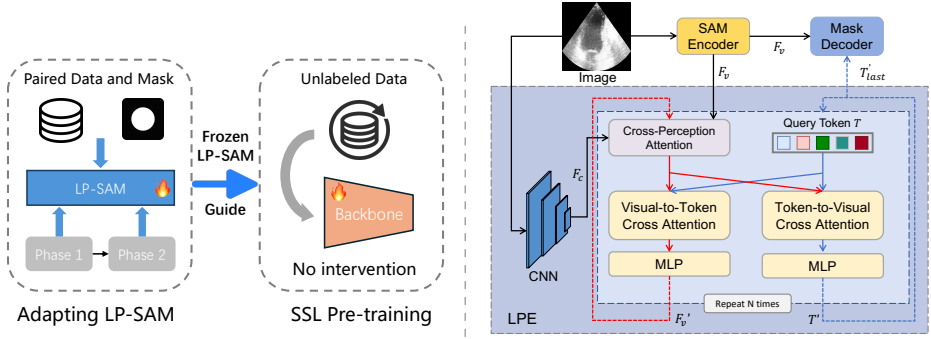

\subsection{Latent-Prompted Domain Adaptation of \med{}}
\label{sec:latsam}

SAM~\cite{kirillov2023segment} exhibits zero-shot segmentation capability on natural images, yet its direct application to US data yields suboptimal masks owing to the domain gap. \med{} (Fig.~\ref{fig:latent_prompt_engine}) resolves this via a \textit{\text{two-stage lightweight adaptation}} that preserves SAM's capacity while injecting domain-specific inductive biases.

\vspace{1.5mm}
\noindent\textit{Stage 1: Adapter-based domain specialization.}
Following~\cite{wu2023medical}, we insert learnable adapters into each ViT block of SAM's image encoder while keeping the pre-trained weights frozen.
Each adapter implements a bottleneck projection $\phi{:}\,\mathbb{R}^{d}{\to}\mathbb{R}^{d'}$ ($d'{\ll}d$) with a residual bypass, trained on existing US image--mask pairs~\cite{pedraza2015open,wunderling2017comparison,leclerc2019deep} via $\mathcal{L}{=}(1{-}\alpha)\mathcal{L}_{\mathrm{BCE}} + \alpha\mathcal{L}_{\mathrm{Dice}}$. The Dice term proves critical for small-structure US targets with severe foreground--background imbalance.


\vspace{1.5mm}
\noindent\textit{Stage 2: Latent prompt engine.}
Annotation-free deployment demands prompt-free mask generation. Let $F_v \in \mathbb{R}^{c \times h \times w}$ and $F_c \in \mathbb{R}^{c \times h \times w}$ denote the global ViT feature and local CNN feature, respectively. A Cross-Perception Attention (CPA) module fuses complementary representations as:
\begin{equation}
    F_{vc} = \mathit{LN}\!\left(\mathit{MHCA}(F_v,\, F_c,\, F_c) + F_v\right),
\end{equation}
where $\mathrm{MHCA}(\cdot)$ is multi-head cross-attention~\cite{vaswani2017attention} with $F_v$ as query and $F_c$ as key and value. A learnable query token $T \in \mathbb{R}^{k \times c}$ is then iteratively refined over $N$ rounds of bidirectional cross-attention:
\begin{align}
    F_v' &= \mathit{MLP}\!\left(\mathit{LN}\!\left(\mathit{MHCA}(F_{vc},\, T,\, T) + F_{vc}\right)\right), \\
    T'   &= \mathit{MLP}\!\left(\mathit{LN}\!\left(\mathit{MHCA}(T,\, F_{vc},\, F_{vc}) + T\right)\right),
\end{align}
where each round feeds the updated $(F_v', T')$ and original $F_c$ as new inputs. The final token $T'_\mathrm{last}$ serves as the task-conditioned prompt embedding for SAM's mask decoder, enabling fully annotation-free anatomical mask generation. During Phase 2, SAM and its adapters are frozen; both stages are \textit{one-time} procedures that do not participate in subsequent SSL pre-training.

\subsection{View-invariant Anatomical Contrastive Matching}
\label{sec:alc}

Previous SSL methods treat entire images or random crops as the unit of comparison, mixing diagnostically relevant anatomy with noisy background. \ours{} operates at the level of anatomical structures identified by \med{}, ensuring that representation learning is grounded in clinically meaningful semantics.

\vspace{2mm}
\noindent\textit{View construction and mask alignment.}
Given an image $x$, two views $\mathbf{v}_1 = t(x)$ and $\mathbf{v}_2 = t'(x)$ are generated by sampling augmentations $t \sim \mathcal{T}$, $t' \sim \mathcal{T}'$.
The corresponding anatomical masks are transformed identically, yielding aligned mask sets $m$ and $m'$.
From each set, $K$ masks are sampled to form a balanced batch of anatomical instances.

\vspace{2mm}
\noindent\textit{Multi-scale mask pooling.}
Each view is encoded by a dual-branch shared backbone $f_\theta$, producing feature maps $\{f_s\}$ at $S$ spatial scales.
For each mask $m_s$ downsampled to scale $s$, a mask-pooled hidden vector is computed as:
\begin{equation}\label{eq:maskpool}
\small
    h_s = \frac{\sum_{i,j} m_s[i,j] \cdot f_s[i,j]}{\sum_{i,j} m_s[i,j]},
\end{equation}
computing the area-normalized mean feature within the region of interest, eliminating background contamination. This approach is more targeted than the heuristic-driven region proposals used in prior object-centric methods~\cite{henaff2021efficient,wen2022self}, which are ill-suited to the low-contrast boundaries common in US images.

\vspace{2mm}
\noindent\textit{Contrastive objective.}
We adopt an asymmetric online--target architecture following BYOL~\cite{grill2020bootstrap}.
The online network maps $h_s$ to a projection $z_{\theta,s} = g_{\theta,s}(h_s)$ and subsequently to a prediction $q_{\theta,s}(z_{\theta,s})$; the target network produces $z'_{\xi,s} = g_{\xi,s}(h'_s)$ with parameters $\xi$ updated via exponential moving average.
After $\ell_2$-normalization $\bar{q}_{\theta,s}{=}q_{\theta,s}/\|q_{\theta,s}\|_2$ and $\bar{z}'_{\xi,s}{=}z'_{\xi,s}/\|z'_{\xi,s}\|_2$,
the separating loss encourages matching anatomy across views while repelling distinct structures:
\begin{equation}
\small
    \ell^{(s)}(\theta;\xi) = -\log \frac{\exp\!\left(\langle \bar{q}_{\theta,s},\, \bar{z}'_{\xi,s} \rangle / \tau\right)}
    {\exp\!\left(\langle \bar{q}_{\theta,s},\, \bar{z}'_{\xi,s} \rangle / \tau\right) + \sum_{n \neq \text{pos}} \exp\!\left(\langle \bar{q}_{\theta,s},\, \bar{z}'_{n,s} \rangle / \tau\right)},
\end{equation}
where $\tau$ is the temperature and the summation over $n$ ranges over all non-matching anatomy instances in the batch.
The total anatomy-level contrastive loss
$
    \mathcal{L}_{\text{ana}}(\theta;\xi) = \frac{1}{K} \sum_{k=1}^{K} \sum_{s=1}^{S} \left( \ell^{(s)}(\theta;\xi) + \tilde{\ell}^{(s)}(\theta;\xi) \right)
$ symmetrises views and aggregates across scales,
where $\tilde{\ell}^{(s)}$ denotes the symmetric counterpart obtained by swapping the roles of views.
Operating across $S$ scales allows the model to capture anatomy at both coarse structural and fine textural granularity.

\subsection{Contextual Prediction over Anatomical Regions}
\label{sec:cep}
Complementing $\mathcal{L}_{\mathrm{ana}}$ with a reconstruction objective over corrupted anatomical regions forces $f_\theta$ to capture structural context beyond surface-level appearance~\cite{rao1999predictive,he2022masked,xie2022simmim}. Unlike random masking~\cite{he2022masked}, our approach targets the \textit{core anatomical region}, \textit{i.e.}, the overlapping crop between $\mathbf{v}_1$ and $\mathbf{v}_2$, ensuring recovery of clinically meaningful content rather than arbitrary background.

Specifically, we extract the overlapping crop shared by $\mathbf{v}_1$ and $\mathbf{v}_2$, partition it into patches, and apply a damage function $D(I; \alpha, \beta, \gamma)$ that combines three perturbation types tailored to US artefacts:
\begin{equation}
    D(x;\,\alpha,\beta,\gamma) = \bigl(\mathcal{O}(\cdot,\gamma) \circ \mathcal{N}(\cdot,\beta) \circ \mathrm{Shuff}(\cdot,\alpha)\bigr)(x),
\end{equation}
where $\mathrm{Shuff}(\cdot, \alpha)$ performs local pixel shuffling with probability $\alpha$ to simulate tissue displacement under probe pressure, $\mathcal{N}(0, \beta)$ injects Gaussian noise mimicking speckle interference, and $\mathcal{O}(\gamma)$ applies random occlusion to emulate acoustic shadowing.
Finally, the $f_\theta$ processes the damaged crop to obtain features $\hat{\mathbf{t}}_{o,s}$, while the $f_\xi$ processes the intact crop to yield $\mathbf{t}_{o,s}$.
The loss is defined as:
   $ \mathcal{L}_{\text{ctx}}(\theta;\xi) = \sum_{s=1}^{S} \|\hat{\mathbf{t}}_{o,s} - \mathbf{t}_{o,s}\|_2^2.$
The pre-training loss integrates both objectives:
\begin{equation}\label{eq:total}
\small
    \mathcal{L}_{\text{total}}(\theta;\xi) = \mathcal{L}_{\text{ana}}(\theta;\xi) + \lambda_{\text{ctx}} \cdot \mathcal{L}_{\text{ctx}}(\theta;\xi),
\end{equation}
where $\lambda_{\text{ctx}}$ controls the relative contribution of contextual prediction.
After pre-training, only the online encoder $f_\theta$ is retained for downstream evaluation; all projection, prediction, and target components are discarded.

\begin{table}[t]
\centering
\large
\caption{\textbf{Performance comparison.} Best in \textbf{bold}, second-best \underline{underlined}. $\dagger$: \med{} masks; $++$: expanded pre-training on \texttt{BF+CM} dataset. POCUS and BUSI are classification benchmarks (accuracy and F1, \%); UDIAT-B, TN3K, BUSI, and HMC-QU are segmentation benchmarks (PPV/Sensitivity/Dice in \%, HD). All differences vs.\ \ours{} on UDIAT-B are significant at $p{<}0.01$.}
\vspace{2mm}
\label{tab:main}
\resizebox{\textwidth}{!}{%
\begin{tabular}{l l c | c c c c | c c | c c c | c c | c c | c c}
\toprule[1.3pt]
& & &
\multicolumn{4}{c|}{\textbf{POCUS}} &
\multicolumn{2}{c|}{\textbf{BUSI}} &
\multicolumn{3}{c|}{\textbf{UDIAT-B}} &
\multicolumn{2}{c|}{\textbf{TN3K}} &
\multicolumn{2}{c|}{\textbf{BUSI}} &
\multicolumn{2}{c}{\textbf{HMC-QU}} \\
\cmidrule(lr){4-7}\cmidrule(lr){8-9}\cmidrule(lr){10-12}\cmidrule(lr){13-14}\cmidrule(lr){15-16}\cmidrule(lr){17-18}
& \textbf{Method} & \textbf{Mask} &
COVID & Pneu. & Reg. & Acc$\uparrow$ &
Acc$\uparrow$ & F1$\uparrow$ &
PPV$\uparrow$ & Sens.$\uparrow$ & Dice$\uparrow$ &
Dice$\uparrow$ & HD$\downarrow$ &
Dice$\uparrow$ & HD$\downarrow$ &
Dice$\uparrow$ & HD$\downarrow$ \\
\midrule[1.3pt]
\multirow{4}{*}{\rotatebox{90}{\large General}}
& Supervised        & \textemdash & 83.7 & 82.1 & 86.5 & 85.0 & 71.3 & 82.8 & 81.5 & 74.7 & 77.9 & 81.57 & 34.31 & 80.55 & 32.91 & 80.70 & 20.72 \\
& ImageNet~\cite{deng2009imagenet} & \textemdash & 79.5 & 78.6 & 88.6 & 84.2 & 84.9 & 81.8 & 82.0 & 80.8 & 81.4 & 81.72 & 33.78 & 81.03 & 32.64 & 81.50 & 21.08 \\
& SimCLR~\cite{chen2020simple}     & \textemdash & 83.2 & 89.4 & 87.1 & 86.4 & 74.6 & 86.3 & 77.4 & 80.8 & 79.1 & 82.10 & 33.25 & 81.12 & 32.45 & 82.14 & 20.49 \\
& MoCo~v2~\cite{chen2020improved}  & \textemdash & 79.7 & 81.4 & 88.9 & 84.8 & 77.8 & 82.8 & 72.6 & 78.6 & 75.4 & 82.17 & 33.54 & 80.13 & 32.59 & 81.42 & 19.96 \\
\midrule[1.3pt]
\multirow{6}{*}{\rotatebox{90}{\large Object-Centric}}
& SlotCon~\cite{wen2022self}        & SG      & 82.2 & 80.9 & 87.3 & 82.8 & 74.1 & 80.1 & 85.3 & 79.9 & 82.7 & 82.02 & 32.15 & 80.28 & 31.74 & 81.51 & 21.23 \\
& SlotCon$\dagger$                  & \med{}  & \underline{91.8} & 88.3 & 90.1 & 89.2 & 78.9 & 89.5 & 88.1 & 81.4 & 84.0 & 84.12 & 28.67 & \underline{82.50} & 31.55 & 80.11 & 21.05 \\
& DetCon~\cite{henaff2021efficient} & MCG     & 81.9 & 80.4 & 87.1 & 82.4 & 70.1 & 79.5 & 84.0 & 79.3 & 81.1 & 81.85 & 32.40 & 80.12 & 32.90 & 80.07 & 21.11 \\
& DetCon$\dagger$                   & \med{}  & 86.2 & 89.4 & 91.0 & 88.1 & 82.3 & 91.9 & 87.2 & \underline{81.7} & 83.4 & 83.16 & 31.75 & 82.42 & 31.80 & 82.05 & 20.10 \\
& R2O~\cite{gokul2022refine}        & K-means & 82.3 & 81.7 & 88.3 & 84.5 & 72.8 & 83.5 & 82.7 & 80.5 & 81.6 & 81.65 & 32.62 & 81.78 & 31.68 & 81.16 & 21.10 \\
& R2O$\dagger$                      & \med{}  & 89.5 & 91.2 & 91.8 & 90.9 & 79.3 & 88.9 & 83.6 & 81.2 & 82.2 & 83.70 & 29.27 & 82.24 & 31.42 & 82.21 & 20.82 \\
\midrule[1.3pt]
\multirow{5}{*}{\rotatebox{90}{\large US-Specific}}
& USCL$++$~\cite{chen2021uscl}           & \textemdash & 86.3 & 90.4 & 93.4 & 90.9 & \underline{85.6} & \underline{93.2} & 88.2 & 80.5 & 84.4 & 84.01 & 29.25 & 82.12 & 31.45 & 82.64 & 20.49 \\
& USF-MAE$++$~\cite{megahed2025usf}  & \textemdash & 90.1 & 92.8 & \underline{94.0} & \underline{93.2} & \underline{85.6} & \underline{93.2} & 88.6 & 81.1 & \underline{84.5} & \underline{84.20} & 29.70 & 82.30 & 31.35 & 82.48 & 20.22 \\
& USUCL$++$~\cite{basu2022unsupervised}  & \textemdash & 89.5 & 94.9 & 93.3 & 92.3 & 84.2 & 93.0 & 88.4 & 81.2 & 84.1 & 83.50 & 28.98 & 82.35 & 31.60 & 82.05 & 20.35 \\
& AWCL$++$~\cite{fu2022anatomy}    & \textemdash      & 88.3 & 87.6 & 91.5 & 91.7 & 84.5 & 91.2 & 87.4 & 80.9 & 83.8 & 83.42 & 30.15 & 81.78 & 31.90 & 81.95 & 20.65 \\
& IVPP$++$~\cite{vanberlo2024intra}      & \textemdash & 91.5 & \underline{95.1} & 93.2 & \underline{93.2} & 83.5 & 92.5 & \underline{89.1} & 80.3 & 83.6 & 83.85 & \underline{28.66} & 81.85 & \underline{31.06} & \underline{82.65} & \underline{19.88} \\
\midrule[1.3pt]
& \cellcolor{gray!12}\textbf{\ours{} (ours)} & \cellcolor{gray!12}\med{} &
\cellcolor{gray!12}\textbf{93.4} & \cellcolor{gray!12}\textbf{95.4} & \cellcolor{gray!12}\textbf{96.1} & \cellcolor{gray!12}\textbf{96.2} &
\cellcolor{gray!12}\textbf{87.1} & \cellcolor{gray!12}\textbf{94.3} &
\cellcolor{gray!12}\textbf{92.7} & \cellcolor{gray!12}\textbf{82.3} & \cellcolor{gray!12}\textbf{85.9} &
\cellcolor{gray!12}\textbf{85.42} & \cellcolor{gray!12}\textbf{26.31} &
\cellcolor{gray!12}\textbf{83.35} & \cellcolor{gray!12}\textbf{30.45} &
\cellcolor{gray!12}\textbf{83.48} & \cellcolor{gray!12}\textbf{19.41} \\
\bottomrule[1.3pt]
\end{tabular}}
\end{table}

\section{Experiments}
\label{sec:experiments}

\subsection{Experimental Setup}

\paragraph{Implementation Details.}
\ours{} is implemented in PyTorch with a ResNet-18 FPN backbone~\cite{lin2017feature} initialized from ImageNet weights~\cite{deng2009imagenet}. The training consists of two phases on two RTX~4090 GPUs. First, \med{} adapter fine-tuning runs for 200 epochs (Adam, lr~$=5{\times}10^{-4}$, batch~16), followed by 100 epochs of latent prompt engine training with SAM frozen. Second, SSL pre-training proceeds for 300 epochs via LARS with a cosine-decay schedule and three-epoch warm-up (base lr~$=0.3$, batch~192). Key hyperparameters are set as follows: $\tau=0.1$, $K=16$, $S=4$, $\lambda_{\mathrm{ctx}}=1$, $N=4$, and damage parameters $(\alpha,\beta,\gamma)=(0.15,0.2,0.05)$.

\paragraph{Datasets, Baselines, and Evaluation Protocol.}
We utilize distinct datasets across different stages. For \med{} fine-tuning, we use DDTI~\cite{pedraza2015open} (637 images), TG3K~\cite{wunderling2017comparison} (3,585 images), and CAMUS~\cite{leclerc2019deep} (57,696 images). SSL pre-training is conducted on a combined dataset (denoted \texttt{BF+CM}) comprising Butterfly~\cite{butterfly2020} and CAMUS~\cite{leclerc2019deep}. Downstream evaluation spans \textbf{classification} on POCUS~\cite{born2021accelerating} and BUSI~\cite{al2020dataset} (evaluated via accuracy and F1), and \textbf{segmentation} on UDIAT-B~\cite{yap2020breast}, TN3K~\cite{gong2023thyroid}, BUSI, and HMC-QU (PPV, Sensitivity, Dice, and Hausdorff distance). All downstream tasks employ five-fold cross-validation. We benchmark against three categories: \textit{(i)}~\textbf{general SSL} (ImageNet~\cite{deng2009imagenet}, SimCLR~\cite{chen2020simple}, FixMatch~\cite{sohn2020fixmatch}, MoCo~v2~\cite{chen2020improved}); \textit{(ii)}~\textbf{object-centric methods} (SlotCon~\cite{wen2022self}, DetCon~\cite{henaff2021efficient}, R2O~\cite{gokul2022refine}), with $\dagger$ denoting \med{} mask substitution; and \textit{(iii)}~\textbf{US-specific pre-training methods} (USCL~\cite{chen2021uscl}, USF-MAE~\cite{megahed2025usf}, USUCL~\cite{basu2022unsupervised}, AWCL ~\cite{fu2022anatomy}, IVPP~\cite{vanberlo2024intra}), with $++$ denoting models pre-trained on the \texttt{BF+CM} set.

\begin{figure*}[t]
    \centering
    \includegraphics[width=0.99\linewidth]{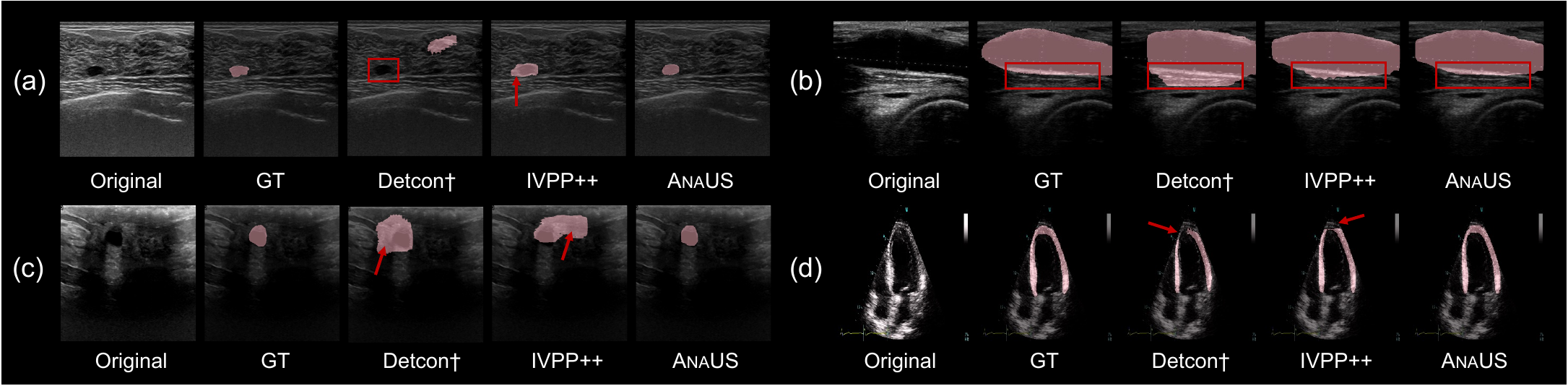}
    \caption{Qualitative downstream segmentation comparison on UDIAT-B, TN3K, BUSI, and HMC-QU. Left to right: Input, ground truth (GT), baselines, \ours{}. 
    }
    \label{fig:segment}
\end{figure*}

\begin{figure}[t]
    \centering
    \includegraphics[width=0.6\linewidth]{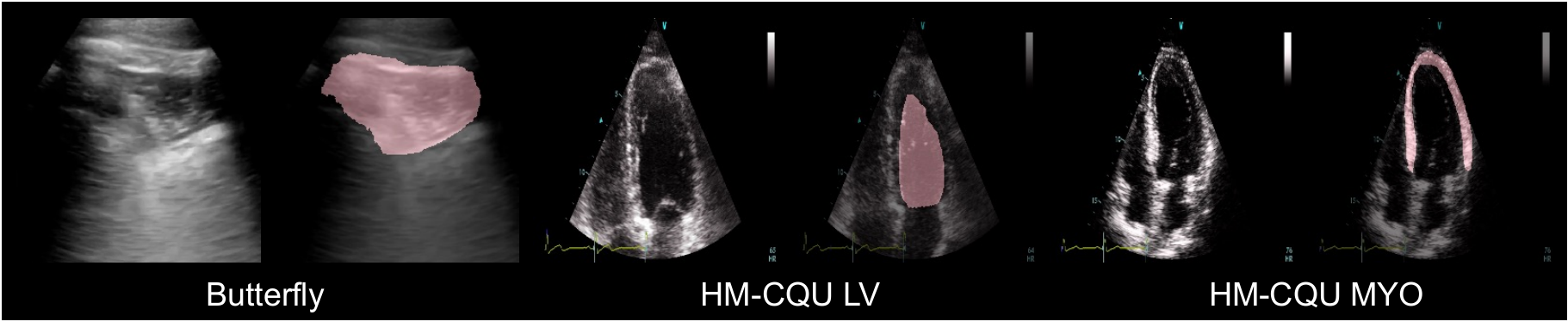}
\caption{Zero-shot \med{} generalization of out-of-distribution structures (bat sign, LA, MYO). The latent $T'_\mathrm{last}$ enables domain-agnostic anatomical discovery.}
    \vspace{-4mm}
    \label{fig:butter_seg}
\end{figure}

\subsection{Comparison with State-of-the-Art Methods}

As shown in Table~\ref{tab:main}, \ours{} achieves the best performance across benchmarks.

\inlinesec{Classification.}
\ours{} attains $\mathbf{96.2\%}$ accuracy on POCUS and $\mathbf{87.1\%}$ on BUSI, surpassing the strongest US-specific baseline IVPP$++$ by $\mathbf{+3.0\%}$ and $\mathbf{+3.6\%}$, respectively. Per-class POCUS margins over the supervised baseline reach $+9.7\%$ (COVID-19), $+13.3\%$ (Pneumonia), and $+9.6\%$ (Regular), highlighting the adaptability of anatomy-level SSL to pathology-discriminative tasks.

\inlinesec{Segmentation.}
On UDIAT-B, \ours{} reaches $\mathbf{85.9\%}$ Dice with $\mathbf{92.7\%}$ PPV, with consistent HD reductions across TN3K, BUSI, and HMC-QU. A critical observation is the \textit{\textbf{mask-source effect}}: replacing generic mask sources (multiscale combinatorial grouping/MCG, semantic grouping/SG, and K-means) with \med{} substantially elevates object-centric baselines across all benchmarks ($+2.3\%$ Dice for DetCon, $+1.3\%$ for SlotCon on UDIAT-B), confirming that \textit{\textbf{anatomically precise masks are the decisive factor}} for US pre-training quality rather than the contrastive objective formulation alone.

\inlinesec{Visual Results.}
Fig.~\ref{fig:segment} qualitatively corroborates \ours{}'s superior boundary precision. While DetCon$\dagger$ shows coarse localization and IVPP$++$ over-smooths challenging structures (\textit{e.g.}, thyroid nodules in TN3K and myocardial edges in HMC-QU), \ours{} maintains sharp delineation. Furthermore, Fig.~\ref{fig:butter_seg} confirms \med{}'s zero-shot generalization to unseen anatomy, demonstrating that iterative cross-attention (Sec.~\ref{sec:latsam}) produces structurally generalizable prompt embeddings $T'_\mathrm{last}$ rather than dataset-specific pattern matching.

\begin{table}[t]
    \centering
    \caption{%
        Ablation analysis of \ours{} on UDIAT-B (PPV/Sensitivity/Dice, \%).
        \textit{Left}: component analysis ($\tilde{\ell}$: symmetric loss, LPE: latent prompt engine, MG: multi-scale contrast, CP: contextual prediction).
        \textit{Right}: CEP function decomposition.
        Statistical significance vs.\ full model (paired $t$-test):
        $^{\dagger}$~$p{<}0.05$;\;$^{\ddagger}$~$p{<}0.01$.
    }
    \vspace{2mm}
    \label{tab:ablation_combined}
    \renewcommand{\arraystretch}{0.95}
    %
    \begin{minipage}[t]{0.46\linewidth}
        \centering
        \resizebox{\linewidth}{!}{%
        \begin{tabular}{>{\centering\arraybackslash}p{0.85cm}>{\centering\arraybackslash}p{0.85cm}>{\centering\arraybackslash}p{0.85cm}>{\centering\arraybackslash}p{0.85cm}|>{\centering\arraybackslash}p{1cm} >{\centering\arraybackslash}p{1.3cm} >{\centering\arraybackslash}p{0.9cm}}
        \toprule[1.3pt]
           $\tilde{\ell}$ & LPE & MG & CP &\textbf{PPV} & \textbf{Sens.} & \textbf{Dice} \\
        \midrule[1.0pt]
        \checkmark &            &  &      \checkmark      & 77.0\rlap{$^{\dagger}$} & 72.3\rlap{$^{\ddagger}$} & 74.3\rlap{$^{\dagger}$} \\
                   & \checkmark & \checkmark &            & 88.5\rlap{$^{\dagger}$}  & 78.5\rlap{$^{\dagger}$}  & 81.2\rlap{$^{\dagger}$}  \\
         & \checkmark &      \checkmark      &       \checkmark     & 84.4\rlap{$^{\ddagger}$} & 74.3\rlap{$^{\ddagger}$} & 77.6\rlap{$^{\dagger}$} \\
        \checkmark &            &  \checkmark & \checkmark & 72.4\rlap{$^{\dagger}$} & 68.1\rlap{$^{\dagger}$} & 70.2\rlap{$^{\ddagger}$} \\
        \checkmark & \checkmark &            & \checkmark & 89.5\rlap{$^{\dagger}$}  & 80.5\rlap{$^{\dagger}$}  & 83.8\rlap{$^{\dagger}$}  \\
        \checkmark & \checkmark & \checkmark &            & 91.1\rlap{$^{\dagger}$}  & 81.6\rlap{$^{\dagger}$}  & 84.9\rlap{$^{\dagger}$}  \\
        \rowcolor{gray!10}
        \checkmark & \checkmark & \checkmark & \checkmark & \textbf{92.7} & \textbf{82.3} & \textbf{85.9} \\
        \bottomrule[1.3pt]
        \end{tabular}}
    \end{minipage}%
    \hfill
    \begin{minipage}[t]{0.48\linewidth}
        \centering
        \resizebox{\linewidth}{!}{%
        \begin{tabular}{p{4cm} >{\centering\arraybackslash}p{1cm} >{\centering\arraybackslash}p{1.1cm} >{\centering\arraybackslash}p{0.9cm}}
        \toprule[1.3pt]
        \textbf{Damage Component} & \textbf{PPV} & \textbf{Sens.} & \textbf{Dice} \\
        \midrule[1.0pt]
        \rowcolor{gray!10}
        \textbf{\ours{} (Full)}                    & \textbf{92.7} & \textbf{82.3} & \textbf{85.9} \\
        \midrule
        \textit{w/o} $\mathrm{Shuff}(\cdot,\alpha)$ & 91.8\rlap{$^{\dagger}$} & 82.0\rlap{$^{\dagger}$} & 84.9\rlap{$^{\dagger}$} \\
        \textit{w/o} $\mathcal{N}(0,\beta)$         & 92.3\rlap{$^{\dagger}$} & 81.9\rlap{$^{\dagger}$} & 85.4\rlap{$^{\dagger}$} \\
        \textit{w/o} $\mathcal{O}(\gamma)$           & 91.9\rlap{$^{\dagger}$} & 82.2\rlap{$^{\dagger}$} & 85.1\rlap{$^{\dagger}$} \\
        \textit{w/o} CEP entirely                    & 91.1\rlap{$^{\dagger}$} & 81.6\rlap{$^{\dagger}$} & 84.7\rlap{$^{\dagger}$} \\
        \bottomrule[1.3pt]
        \end{tabular}}
    \end{minipage}
\end{table}

\subsection{Comprehensive Ablation Analysis}

\inlinesec{Component Analysis.}
Table~\ref{tab:ablation_combined} (left) dissects each design choice. Substituting \med{} with vanilla SAM (\textit{i.e.}, w/o LPE) incurs the sharpest degradation ($-15.7\%$ Dice), underscoring \textit{\textbf{domain-adapted mask}} as a prerequisite for grounded SSL. Removing symmetric contrast ($\tilde{\ell}$) drops Dice by $8.3\%$ ($p{<}0.01$), while disabling multi-scale contrast (MG) and contextual prediction (CP) further reduces Dice by $2.1\%$ and $1.0\%$ respectively ($p{<}0.05$), with each component proving complementary.

\inlinesec{Damage Function Decomposition.}
Table~\ref{tab:ablation_combined} (right) isolates each perturbation in the CEP damage function. Pixel shuffling yields the largest gain, followed by occlusion and Gaussian noise, with each component simulating distinct US imaging artefacts; removing all three degrades performance by $1.2\%$.

\section{Conclusion}

We present \ours{}, an anatomy-anchored self-supervised pre-training framework that shifts the contrastive unit from arbitrary patches to clinically meaningful structures. \med{} bridges the natural-to-ultrasound domain gap via lightweight adapters and a learnable latent prompt engine, enabling annotation-free anatomical discovery. Inter-view anatomy-level contrast and contextual prediction jointly enforce inter-anatomy discriminability and structural coherence. Extensive evaluations confirm that \ours{} establishes a principled and clinically aligned foundation for future research in medical self-supervised learning.

\begin{credits}
\subsubsection{Acknowledgments.} This research was partially supported by the National Major Scientific Instrument Development Project of the National Natural Science Foundation of China (Grant No. 62227808), the National Natural Science Foundation of China (Grants No. 62472157), and the Hunan Provincial Graduate Research Innovation Project (Grant No. CX20250587).
\subsubsection{Disclosure of Interests.}The authors have no competing interests to declare that are relevant to the content of this article.
\end{credits}


\bibliographystyle{splncs04}
\bibliography{reference}

\clearpage
\newpage

\appendix
%
%
%
%

\setcounter{subsection}{0}
\renewcommand{\thesubsection}{A.\arabic{subsection}}
\subsection{Supplementary Experimental Details}
\label{sec:supp_details}

\inlinesec{Data augmentation.}
Both views follow the BYOL protocol: random crop ($0.2$--$1.0$), horizontal flip, colour jitter, Gaussian blur, and grayscale; $\mathbf{v}_2$ additionally applies solarisation. Geometric transforms are applied identically to masks.

\inlinesec{Downstream fine-tuning.}
For \textbf{classification} (POCUS, BUSI), only the last three backbone layers are fine-tuned using Adam (lr~$=0.01$) for 30 epochs.
For \textbf{segmentation} (UDIAT-B, TN3K, BUSI, HMC-QU), we employ Mask-RCNN with the pre-trained backbone, optimised via SGD (weight decay~$10^{-4}$).

\inlinesec{Mask sampling.}
Each binary mask is downsampled via adaptive average pooling to match backbone feature maps at $\{7{\times}7,\;14{\times}14,\;28{\times}28,\;56{\times}56\}$.
Weighted by spatial coverage, $K$ masks are drawn from the resulting categorical distribution, ensuring anatomically salient structures receive sampling priority.

\inlinesec{Computational cost.}
\med{} adapter fine-tuning introduces \textbf{13M} parameters and requires ${\sim}$24\,h; the latent prompt engine adds \textbf{11.1M} parameters (${\sim}$8\,h). Full \ours{} pre-training (300 epochs, batch 192) completes in ${\sim}$10\,h on two RTX~4090 GPUs (${\sim}$22\,GB VRAM). Both \med{} stages are \textit{one-time} procedures excluded from SSL iterations. The retained encoder processes a $224{\times}224$ image in ${\sim}$\textbf{0.45\,ms} on a RTX~4090, confirming \textit{\textbf{real-time clinical viability}}.

%
\begin{table}[t]
\centering
\caption{Hyperparameter sensitivity on UDIAT-B (Dice, \%). Default settings are \textbf{bolded}; $\uparrow$ marks peak performance.}
\label{tab:supp_hyper}
\small
\setlength{\tabcolsep}{3.5pt} 
\resizebox{0.85\textwidth}{!}{%
\begin{tabular}{cc|cc|cc|cc|cc}
\toprule
Temp. $\tau$ & Dice & Scales $S$ & Dice & $\lambda_{\mathrm{ctx}}$ & Dice & $N$ (LPE) & Dice & Prob. $\alpha$ & Dice \\
\cmidrule{1-2}\cmidrule{3-4}\cmidrule{5-6}\cmidrule{7-8}\cmidrule{9-10}
0.05 & 84.8 & 1 & 79.8 & 0.1 & 84.5 & 2 & 84.6 & 0.05 & 85.2 \\
\textbf{0.10} & \textbf{85.9}$\uparrow$ & 2 & 83.0 & 0.5 & 85.3 & \textbf{4} & \textbf{85.9}$\uparrow$ & 0.10 & 85.6 \\
0.20 & 85.2 & 3 & 84.6 & \textbf{1.0} & \textbf{85.9}$\uparrow$ & 6 & 85.7 & \textbf{0.15} & \textbf{85.9}$\uparrow$ \\
0.50 & 84.1 & \textbf{4} & \textbf{85.9}$\uparrow$ & 2.0 & 85.0 & 8 & 85.5 & 0.20 & 85.4 \\
\bottomrule
\end{tabular}
}
\end{table}

\subsection{Hyperparameter Sensitivity}
\label{sec:supp_hyper}

Table~\ref{tab:supp_hyper} reports UDIAT-B Dice under systematic variation of key hyperparameters.
$\tau{=}0.1$ sharpens \textit{\textbf{inter-anatomy separability}} without over-penalising hard negatives; $\lambda_{\mathrm{ctx}}{=}1.0$ and $N{=}4$ cross-attention rounds yield stable peaks, with diminishing returns beyond these values.
For the Stage~1 adapter loss, the combined formulation $(1{-}\alpha)\mathcal{L}_{\mathrm{BCE}} + \alpha\mathcal{L}_{\mathrm{Dice}}$ improves mask IoU, confirming that the Dice term is \text{\text{critical for foreground--background imbalance}} in US images.


\begin{figure}[htpb]
    \centering
    \resizebox{\textwidth}{!}{%
        \begin{minipage}[t]{0.33\textwidth}
            \centering
            \includegraphics[width=\textwidth]{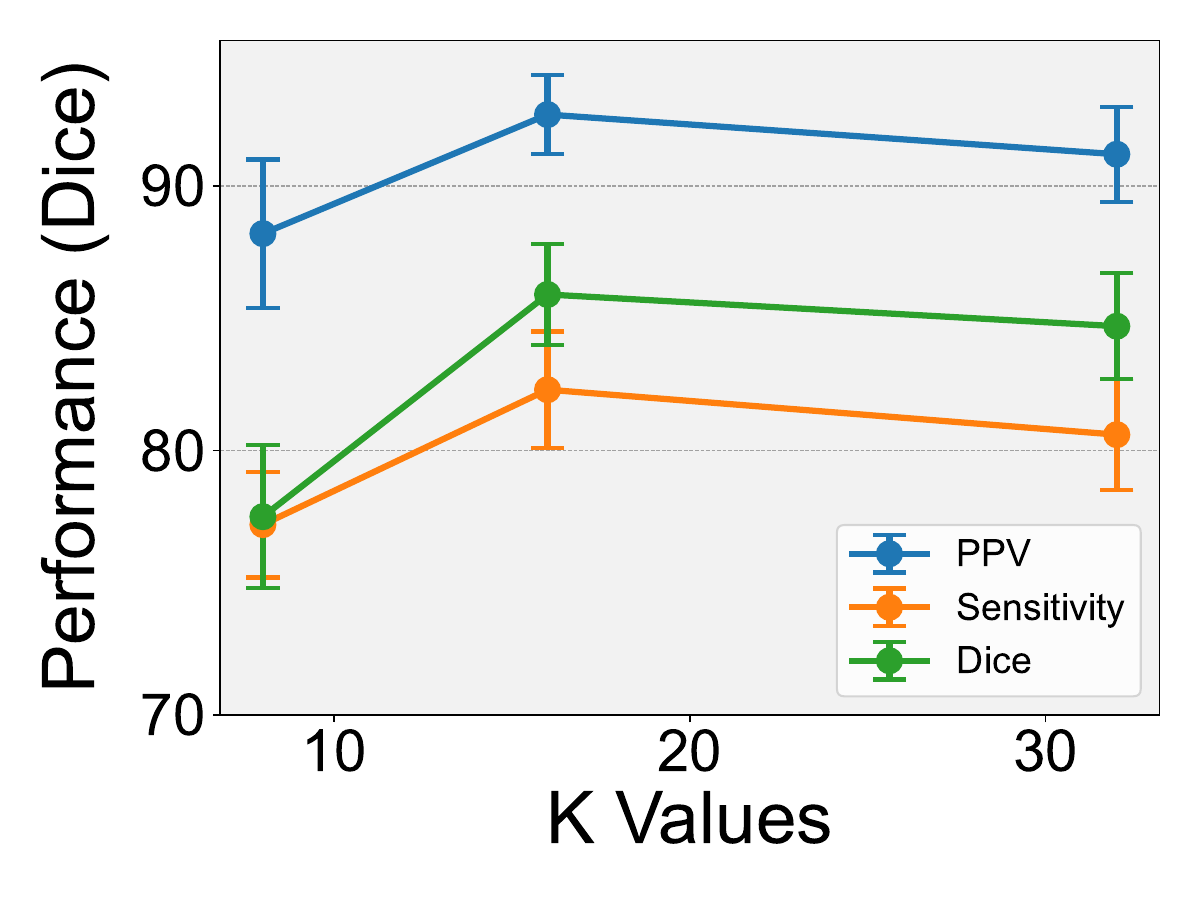}
        \end{minipage}
        \hfill
        \begin{minipage}[t]{0.33\textwidth}
            \centering
            \includegraphics[width=\textwidth]{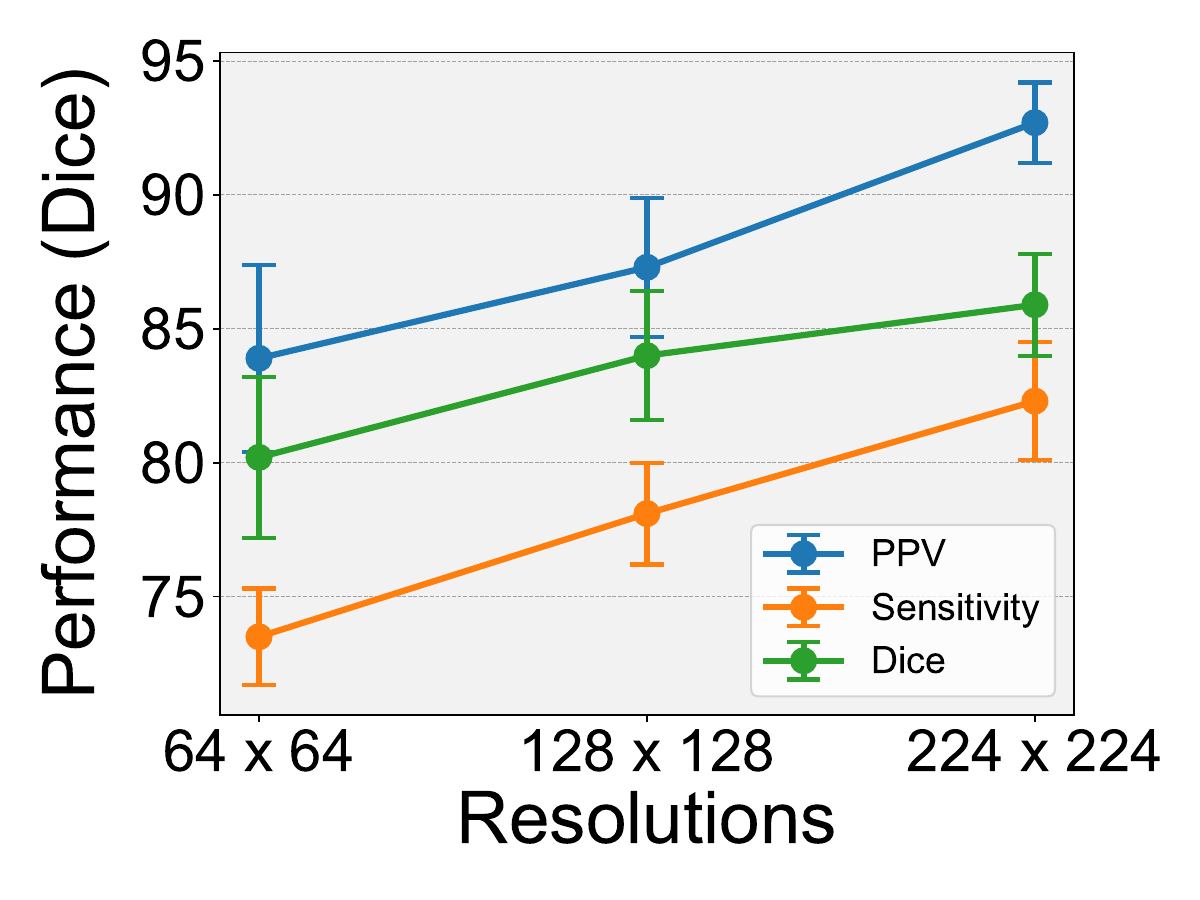}
        \end{minipage}
        \hfill
        \begin{minipage}[t]{0.33\textwidth}
            \centering
            \includegraphics[width=\textwidth]{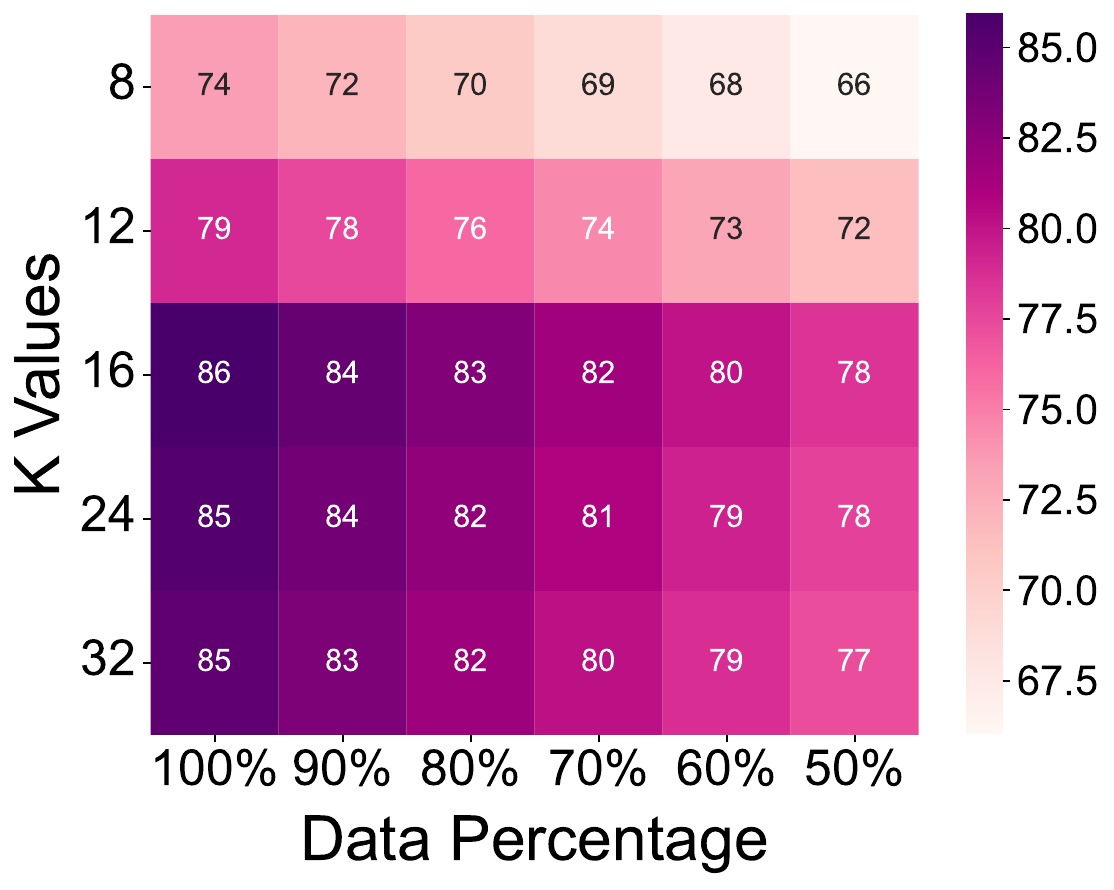}
        \end{minipage}
    }
    \caption{Ablation on UDIAT-B. \textit{(a)} Effect of mask sample count $K$. \textit{(b)} Pre-training resolution. \textit{(c)} Joint effect of $K$ and data percentage on Dice.}
    \label{fig:supp_ablation}
\end{figure}

\subsection{Pre-training Resolution, Data Volume, and $K$}
\label{sec:supp_resolution}

Fig.~\ref{fig:supp_ablation} presents three complementary analyses.
\textit{(a)}~$K{=}16$ achieves the best PPV/Sensitivity/Dice balance; $K{=}32$ improves PPV but degrades Sensitivity, suggesting \textit{\textbf{redundant masks introduce noise}}.
\textit{(b)}~Resolution $224{\times}224$ preserves fine-grained anatomical detail essential for boundary delineation; 
\textit{(c)}~The heatmap confirms the joint optimum at $K{=}16$ with 100\% data; performance degrades notably below 60\% data under suboptimal $K$, underscoring the complementary roles of data volume and mask sampling.

\begin{figure}[t]
\begin{minipage}[t]{0.52\linewidth}
\vspace{0pt}
\captionof{table}{\ours{} vs.\ IVPP$++$ across backbone scales on UDIAT-B.}
\label{tab:supp_scale}
\centering
\small
\setlength{\tabcolsep}{3pt}
\renewcommand{\arraystretch}{1.1}
\resizebox{\linewidth}{!}{%
\begin{tabular}{ll ccc}
\toprule
\textbf{Backbone} & \textbf{Method} & \textbf{PPV$\uparrow$} & \textbf{Sens.$\uparrow$} & \textbf{Dice$\uparrow$} \\
\midrule
\multirow{2}{*}{\makecell{ResNet-18 \\ {\scriptsize (0.45\,ms)}}} & IVPP$++$ & 89.1 & 80.3 & 83.6 \\
 & \ours{} & \textbf{92.7} & \textbf{82.3} & \textbf{85.9} \\
\midrule
\multirow{2}{*}{\makecell{ResNet-50 \\ {\scriptsize (1.12\,ms)}}} & IVPP$++$ & 90.2 & 81.1 & 84.5 \\
 & \ours{} & \textbf{93.5} & \textbf{83.0} & \textbf{86.5} \\
\midrule
\multirow{2}{*}{\makecell{ResNet-101 \\ {\scriptsize (1.84\,ms)}}} & IVPP$++$ & 90.8 & 81.7 & 85.1 \\
 & \ours{} & \textbf{94.1} & \textbf{83.5} & \textbf{87.3} \\
\midrule
\multirow{2}{*}{\makecell{ResNet-152 \\ {\scriptsize (2.53\,ms)}}} & IVPP$++$ & 91.3 & 82.0 & 85.4 \\
 & \ours{} & \textbf{94.6} & \textbf{84.2} & \textbf{87.5} \\
\bottomrule
\end{tabular}}
\end{minipage}%
\hfill
\begin{minipage}[t]{0.44\linewidth}
\vspace{0pt}
\centering
\includegraphics[width=\linewidth]{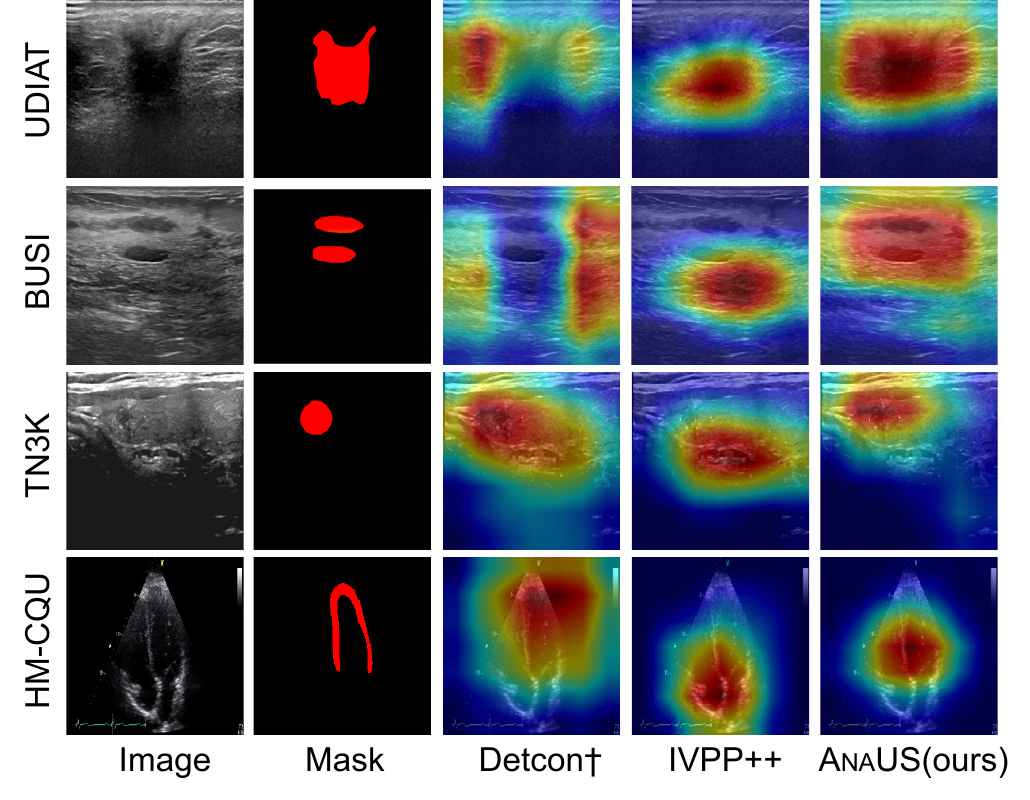}
\captionof{figure}{Grad-CAM visualization. \ours{} focuses on anatomy with minimal background dispersion.}
\label{fig:supp_gradcam}
\end{minipage}
\end{figure}
\vspace{-2mm}

\subsection{Impact of Backbone Scale}
\label{sec:supp_scale}

Table~\ref{tab:supp_scale} compares \ours{} and IVPP$++$ across four ResNet variants on UDIAT-B.
\ours{} consistently outperforms at every scale, with the Dice gap sustained from $+2.3\%$ (ResNet-18) to $+2.1\%$ (ResNet-152).
Notably, \ours{} with ResNet-18 already surpasses IVPP$++$ with ResNet-50, demonstrating that anatomy-anchored pre-training compensates for limited model capacity.

\subsection{Grad-CAM Interpretability}
\label{sec:supp_gradcam}

Fig.~\ref{fig:supp_gradcam} visualizes final-layer activations via Grad-CAM across four datasets.
DetCon$\dagger$ produces diffuse activations extending beyond lesion boundaries; IVPP$++$ improves localization but retains non-negligible background response.
\ours{} generates activations \textit{\textbf{tightly confined to target anatomy}}, confirming that the anatomy-level contrastive objective suppresses background encoding and internalises diagnostically relevant spatial priors and representative features.



\end{document}